\documentclass[runningheads]{llncs}
\usepackage{mathtools}
\usepackage{tabularx}
\usepackage{tablefootnote}
\usepackage[dvipsnames]{xcolor}
\usepackage{multirow}
\usepackage{graphicx}
\usepackage{subcaption}
\usepackage[utf8]{inputenc}
\usepackage{float}
\usepackage[linesnumbered,ruled]{algorithm2e}
\usepackage{caption}

\usepackage[colorlinks,allcolors=blue]{hyperref}

\begin{document}
\title{Targeted Honeyword Generation with Language Models}

\titlerunning{HoneyGAN}

\author{Fangyi Yu\inst{1}\orcidID{0000-0002-6461-5720} \and
Miguel Vargas Martin\inst{1}\orcidID{0000-0001-8169-6836}}

\institute{Ontario Tech University, Oshawa ON L1L 1C1, Canada\\
\email{\{fangyi.yu,Miguel.VargasMartin\}@ontariotechu.ca}}

\maketitle              
\begin{abstract}
Honeywords are fictitious passwords inserted into databases in order to identify password breaches. The major difficulty is how to produce honeywords that are difficult to distinguish from real passwords. Although the generation of honeywords has been widely investigated in the past, the majority of existing research assumes attackers have no knowledge of the users. These honeyword generating techniques (HGTs) may utterly fail if attackers exploit users' personally identifiable information (PII) and the real passwords include users' PII. In this paper, we propose to build a more secure and trustworthy authentication system that employs off-the-shelf pre-trained language models which require no further training on real passwords to produce honeywords while retaining the PII of the associated real password, therefore significantly raising the bar for attackers. 

We conducted a pilot experiment in which individuals are asked to distinguish between authentic passwords and honeywords when the username is provided for GPT-3 and a tweaking technique. Results show that it is extremely difficult to distinguish the real passwords from the artifical ones for both techniques. We speculate that a larger sample size could reveal a significant difference between the two HGT techniques, favouring our proposed approach.
\end{abstract}

\section{Introduction}
Passwords have dominated the authentication system for decades, despite their security flaws compared to competing techniques such as cognitive authentication \cite{hopper2001secure}, biometrics \cite{ratha2001enhancing} and tokens \cite{roche2021side}. The irreplaceable is primarily due to its incomparable deployability and usability \cite{bonneau2012quest}. However, current password-based authentication systems store sensitive password files that make them ideal targets for attackers because if successfully obtained and cracked (recovering the hashed passwords' plain-text representations), an adversary may impersonate registered users in an undetectable fashion \cite{wang2018security}. Numerous prestigious online services have been infiltrated, for example, Yahoo!, RockYou, Zynga, resulting in the exposure of millions of credentials. Unfortunately, there is often a large delay between a credential database's breach and its detection; estimates place the average latency at 287 days \cite{IBM}. The resulting window of vulnerability enables attackers to crack passwords offline (if the stolen credential database contains encrypted passwords rather than plain-text passwords); determine the value of these passwords by probing their associated accounts, and then use them directly to extract value or sell them via illicit forums profiting with stolen credentials \cite{thomas2017data}. Normally, the longer it takes to detect and remediate a data breach, the more expensive it is. More precisely, data breaches that take more than 200 days to identify and contain cost an average of 4.87 million, compared to 3.61 million for breaches that take fewer than 200 days to identify and manage \cite{IBM}. As a result, it is vital to have active, timely password-breach detection systems in place to allow immediate counter-actions.

One way to reduce the cost of password breaches is to make offline guessing harder. A variety of ways have been proposed in the literature, including machine-dependent functions \cite{almeshekah2015ersatzpasswords},  external password-hardening services \cite{lai2017phoenix}, and distributed cryptography \cite{camenisch2015optimal}. All of these approaches, however, have major disadvantages, such as low scalability or a need for large modifications to the server-side and client-side authentication systems, which prevent the community from implementing them.

Another promising approach is to shorten the latency between password breaches and detection. Juels and Rivest suggest the use of honeywords as a potential method for efficiently detecting password leaks \cite{juels2013honeywords}. According to their proposal, a website could store decoy passwords, called honeywords, alongside real passwords in its credential database, so that even if an attacker steals and reverts the password file containing the users' hashed passwords, they must still choose a real password from a set of $k$ distinct \textit{sweetwords}, where a real password and its associated honeywords are referred to as sweetwords. The attacker's use of a honeyword could cause the website to become aware of the breach. Notably, honeywords are only beneficial if they are difficult to distinguish from real-world passwords; otherwise, a knowledgeable attacker may be able to recognize them and compromise their security. Thus, when implementing this security feature into current authentication systems, the honeyword generating process is critical.
\begin{table}[]
\centering
\caption{Data  breaches containing PII and passwords in the past five years}
\label{tab:data_breach}
\resizebox{\textwidth}{!}{%
\begin{tabular}{|c|c|c|c|}
\hline
\textbf{Dataset} & \textbf{Number of Items} & \textbf{Year} & \textbf{Type of PII breached}                                      \\ \hline
Neiman Marcus & 4,800,000 & 2021 & Name, Encrypted Password, Security questions, Financial information \\ \hline
CAM4             & 10,880,000,000           & 2020          & Name, Email, Encrypted Password, Chat transcripts, IP, Payment logs \\ \hline
Canva            & 137,000,000              & 2019          & Name, Email, Encrypted Password                                    \\ \hline
Quora            & 100,000,000              & 2018          & Name, Email, Encrypted Password, Questions and answers posted     \\ \hline
Yahoo            & 3,000,000,000            & 2017          & Name, Email, Encrypted Password, DoB, Security question and answer \\ \hline
\end{tabular}%
}
\end{table}
\subsection{Honeywords for Targeted Attacks} \label{targeted}
The biggest challenge of designing a honeyword generation technique is to generate honeywords that are resistant to targeted attacks. For targeted attacks, attackers exploit users’ PII to guess passwords, which increases the likelihood of users’ accounts being compromised. This is a critical problem for two reasons. First, numerous PII and passwords become widely accessible as a result of ongoing data breaches \cite{pwned,IBM}. Second, people are used to create easy-to-remember passwords using their names, birthdays, and their variants \cite{wang2016targeted}. Once an attacker obtains users’ PII, and if only one sweetword in a user’s sweetword list contains the user’s PII, it is highly likely that this sweetword is the real password and others are fake. For example, for a sweetword list \textit{``liyaodong007, gaby1124, abg71993, australiaisno\#1, soloelbambino, k646321102, noviembre9101, blueluna17, usa0858199600, kirsten03''} which are generated using the real password \textit{``liyaodong007''} in the \textit{linkedin} dataset and the HGT proposed in \cite{dionysiou2021honeygen}. In a nutshell, this HGT is first trained on a real password dataset, and it converts all real passwords in the dataset into vectors using a word embedding technique called \textit{fasttext}. For each user, the HGT assigns $k-1$ honeywords to the $k-1$ real passwords that have the closest distance to this user's actual password based on cosine similarity. In this case, if the attacker has no information about the user, it will be difficult to determine which of the ten sweetwords is the real password, since all of honeywords are from data breaches and are legitimate passwords belonging to other users. However, if the attacker knows the user's email address is \textit{``liyaodong@gmail.com''}, it is quite straightforward to deduce that \textit{``liyaodong007''} is this user's real password and the others are all fake. 

Following the introduction of the honeywords security mechanism by Juels and Rivest \cite{juels2013honeywords}, the academic community has been actively exploring the technique. However, to our knowledge, only one paper \cite{wangattack} concentrated on the production of honeywords in a targeted manner. All other papers make the invalid assumption that attackers have no knowledge about the users. Each year, as demonstrated in Table \ref{tab:data_breach}, billions of password datasets including PII are leaked. Attackers might use the PII to determine which sweetword is the real password. If all the sweetwords do not include any PII existing in the password breach, the attackers may still create a knowledge map for each user by searching their information purposefully through social media and search engines using the known PII exposed in data breaches. This is especially a concern if the user is a celebrity or a politician. Compromised accounts may have substantial financial, political, and societal consequences.

\begin{figure*}[h]
\centering
  \includegraphics[width=0.9\textwidth]{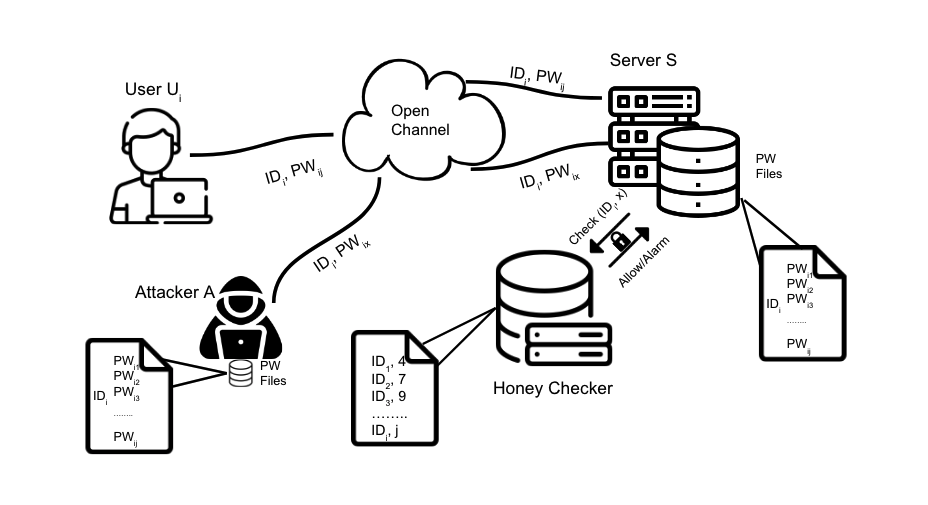}
  \caption{Password (PW) authentication with honeywords.}
  \label{fig:HGT}
\end{figure*} 

\subsection{Related Work}
Numerous studies have been conducted on the non-targeted honeyword generation method. The majority of these HGTs fall into two categories: chaffing-by-tweaking and chaffing-with-a-password-model. Chaffing-by-tweaking is mostly based on the substitution of random letters, digits, and symbols. For instance, given the real password ``$deshaun96$'', we could get honeywords ``$deshaun87$, $deshAUn66$, $DesHaun56$'' via tweaking. However, as Wang et al. \cite{wang2018security} demonstrate, this strategy is indeed vulnerable. While honeywords generated using the chaffing-with-a-password-model approach are more resistant to attacks, they do have certain drawbacks. Bojinov et al. \cite{bojinov2010kamouflage} proposed \textit{Kamouflage}, that first tokenize the user's real passwords into a collection of tokens, and then substitutes each token with a random one that matches the token's type. For instance, ``$jones34monkey$'' is tokenized as ``$l_5d_2l_6$'' (a five-letter word followed by two digits and a six-letter word), indicating that some possible honeywords are ``$apple10laptop$, $tired93braces$,  $hills28highly$''. This technique, as outlined in \cite{dionysiou2021honeygen}, demands considerable modifications on the client-side authentication system, which has a significant impact on usability. Additionally, it is incapable of generating honeywords of varying length or structure, thus limiting the spectrum of possible honeywords.

Erguler \cite{erguler2015achieving} proposed a different technique in which honeywords are derived from the system's current user passwords. In this case, all honeywords are realistic and adhere to the operator's password creating policy. However, their HGT is restricted by the limited number of viable honeywords created by selecting genuine passwords from the website's password corpus, which is particularly the case if the website has a small user database. 

To the best of our knowledge, there is only one publication that discusses how to generate honeywords that are resistant to targeted attacks, which was published in IEEE S\&P’22 by Wang et al \cite{wangattack}. They first proposed four attack models each representing a potential attacker $A$'s strategy, with each model based on different information available to $A$ (e.g., public datasets, the victim's personal information and registration order). They further develop four HGTs for each attack strategy, by using various representative probabilistic password guessing models proposed in their previous paper \cite{wang2016targeted}. These assumptions about attackers are flawed since we should anticipate the attackers would utilize whatever information they can get to attack users' accounts, particularly if the user is a person of interest. What we are proposing is a much simpler and generalized approach. Rather than assuming $A$'s attack strategy and creating HGTs accordingly, we construct honeywords based on the information contained in the real password. The challenge is to partition the real password into tokens while retaining tokens that correspond to PII and replacing tokens that do not correspond to PII with random ones. Consider the real password \textit{``liyaodong007''}, the challenge is to produce honeywords containing the token\textit{``liyaodong''}, which is the user's full name as indicated by his/her email address. To do this, we propose to employ language models, which are extensively used in the area of natural language processing (NLP), to overcome this obstacle.

\subsection{Our contribution}

\begin{itemize}
\item We are among the first to propose HGTs that are resistant to targeted attacks in which attackers have access to users' PII. In comparison to \cite{wangattack}, our solution is much simpler and can be adapted to defend against any form of attack.
\item We are the first to use language models to create honeywords. We suggest building honeywords using GPT-3, which has an easy operator implementation. The off-the-shelf GPT-3 model can create high-quality honeywords that are resistant to targeted attacks without being trained on real passwords. We would like to draw the attention of researchers to use language models to create honeywords that are resistant to targeted attacks, and thus build a more trustworthy authentication system.
\item We conducted a pilot study to establish our method's validity.

\end{itemize}

The remainder of the paper is structured as follows: Section 2 provides the preliminaries for understanding our work. Section 3 introduces our approach to generate honeywords in a targeted manner. Section 4 is the human study design and evaluation. Section 5 discusses the limitations of our work and future directions. Section 6 concludes our work.

\section{Preliminaries}
\subsection{The Honeyword Mechanism} 
According to Juels and Rivest \cite{juels2013honeywords}, the honeyword system is comprised of four entities, as shown in Figure \ref{fig:HGT}: a user $U_i$, an authentication server $S$, a $honeychecker$, and the attacker $A$. User $U_i$ initially registers an account($ID_i$, $PW_i$) on the server $S$. Apart from the standard user registration processes, $S$ runs a command $GEN(k, PW_i)$ to produce a list of $k-1$ unique fake passwords (called honeywords) to be stored alongside $U_i$'s true password $PW_i$, where $k=20$ as recommended in \cite{juels2013honeywords}. $PW_i$ and its $k-1$ honeywords are referred to as $k$ sweetwords.

\subsection{Threat Model}
Honeyword-enabled systems could reliably identify a password file leak by pairing each user's account with $k-1$ honeywords. The reason for this is that even if attackers obtain a copy of the password file along with its hashing parameters and salts, and successfully recover all the passwords via brute-force or other password guessing techniques \cite{durmuth2015omen,weir2009password,melicher2016fast,pasquini2021improving} (be aware that at this stage they know which $k$ sweetwords are associated with each user), they must first distinguish each user's true password from these $k$ sweetwords. The system features \textit{honeychecker} to aid in the usage of honeywords, and the computer system could interact with the \textit{honeychecker} whenever a login attempt is made or users change their passwords. Additionally, the \textit{honeychecker} is capable of triggering an alert if an anomaly is discovered. The warning signal may be sent to an administrator or to a third party other than the computer system itself \cite{juels2013honeywords}.
This approach is compatible with existing authentication systems since it needs little adjustments to the server-side systems and no alterations to the client-side systems; nevertheless, it is very reliable due to the high probability of capturing adversaries. For instance, if the likelihood of an attacker selecting each sweetword is uniform, the probability of capturing an attacker is $3/4 = 75\%$ for $k = 4$, and thus the probability grows as $k$ increases. 

Even though most passwords breached are encrypted, with the evolution of high-performance computing equipment such as GPUs and large-scale distributed clusters, attackers could eventually reverse the majority of password hashes stored in password file \cite{sprengers2011gpu}. As a result, once an attacker obtains any password files, it is reasonable to assume that the overwhelming majority of passwords can be brute-force cracked offline, and converted to their plain-text version.
\begin{table}[]
\centering
\caption{Honeywords generated by GPT-3 when using different prompts. A more specific prompt can produce higher-quality honeywords.}
\label{tab:prompt}
\resizebox{\textwidth}{!}{%
\begin{tabular}{|c|c|}
\hline
\textbf{Prompt1}            & Suggest three passwords that are similar to "toby2009bjs". \\ \hline
\textbf{Honeywords} & toby2009bjd, toby2009bjx, toby2009bjz                      \\ \hline
\textbf{Prompt2}            & Suggest three words that look like "toby2009bjs".          \\ \hline
\textbf{Honeywords} & toy2009bjs, tab2009bjs, boy2009bjs                         \\ \hline
\end{tabular}%
}
\end{table}
\begin{table}[]
\centering
\caption{Honeywords generated by GPT-3 when using different temperatures and given the prompt "Suggest five words that are similar to "toby2009bjs"". A higher temperature will result in more diverse honeywords.}
\label{tab:temp}
\resizebox{\textwidth}{!}{%
\begin{tabular}{|c|c|}
\hline
\textbf{Temperature} & \textbf{Honeywords}                                      \\ \hline
0                    & toby2009bjd, toby2009bjx, toby2009bjz, toby2009bjf, toby2009bjh   \\ \hline
1                    & Toby2009BJS, toby2009bjs1, tobybjs2009, Bjs2009toby, bjs2009toby1 \\ \hline
\end{tabular}%
}
\end{table} 

\subsection{Language Model} Language models can learn the probabilities of occurrences of a series of words in a regularly spoken language (for example, English) and predict the next potential word in that sequence. Generative Pre-trained Transformer 3 (GPT-3) is an autoregressive language model that uses deep learning to generate text that appears to be written by a person. It was introduced in 2020 by Elon Musk et al.'s AI research lab, OpenAI, and excels at a variety of NLP tasks, including translation, question-answering, and cloze \cite{brown2020language}. The model was trained on trillions of words in text documents. It turns the words into vectors, or mathematical representations, and then decode the encoded text into human-readable phrases. It comprehends and processes text by breaking it into tokens. Tokens may be single words or groups of characters. For instance, the word ``beautiful'' could be split into the tokens  ``beau'', ``ti'' and ``ful'', but a short and common word such as ``big'' is a single token.  The model can be utilized to execute NLP tasks without requiring fine-tuning on particular downstream task datasets. It is capable of producing texts that are difficult for humans to differentiate from human-written articles. 

Since the introduction of Generative Pre-trained Transformers, they have been extensively investigated in a variety of domains, including creating summaries of media dialogues \cite{chintagunta2021medically}, generating code from natural-language instructions \cite{chen2021evaluating}, generating passphrases \cite{AiP2021}, and generating graphics from text descriptions \cite{ramesh2021zero}. To the best of our knowledge, we are the first to employ GPT-3 in the sphere of computer security, to generate honeywords that are resistant to targeted attacks.

\section{Approach}
\subsection{Use GPT-3 for targeted honeyword generation}
We propose to use GPT-3 to generate honeywords that are robust to targeted attacks. When generating honeywords, GPT-3 first splits its input, the real password, into tokens. For example, depending on the $temperature$, ``$toby2009bjs$'' may be tokenized to ``$toby$'', ``$2009$'' and ``$b$'', ``$j$'', ``$s$''. When given a $prompt$, GPT-3 generates a text $completion$ that tries to match the context or pattern specified. Since honeyword is a new term specified in the computer security domain and does not exist in the GPT-3's training data, we need to specify what the model should do by giving it a prompt, for example, ``Suggest five passwords that are similar to ``$toby2009bjs$'' .''  It will then produce outputs ``\textit{toby2009bjd, toby2009bjf, toby2009bjg, toby2009bjh, toby2009bjk}.'' The quality and the diversity of the output depends on three attributes: prompt, temperature and examples given to the model.

\textbf{The prompt.} The prompt is the instruction GPT-3 received. The quality of the prompt can determine the quality of the generated honeywords. By experiments, we found that the more concise, specific the prompt is, the higher the quality of the honeywords, as shown in Table \ref{tab:prompt}.

\begin{figure*}
\centering
\begin{subfigure}{1\textwidth}
  \centering
  \includegraphics[width=1\linewidth]{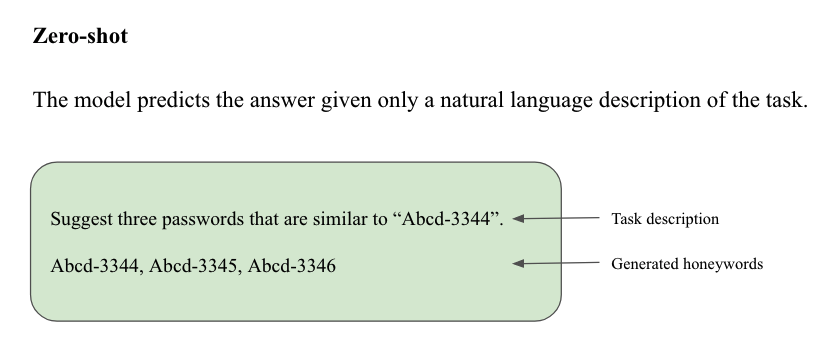}
  \caption{Give GPT-3 no examples when generating honeywords. The honeywords simply adds on the last digit.}
  \label{fig:zero-shot}
\end{subfigure}%
\\
\begin{subfigure}{1\textwidth}
  \centering
  \includegraphics[width=1\linewidth]{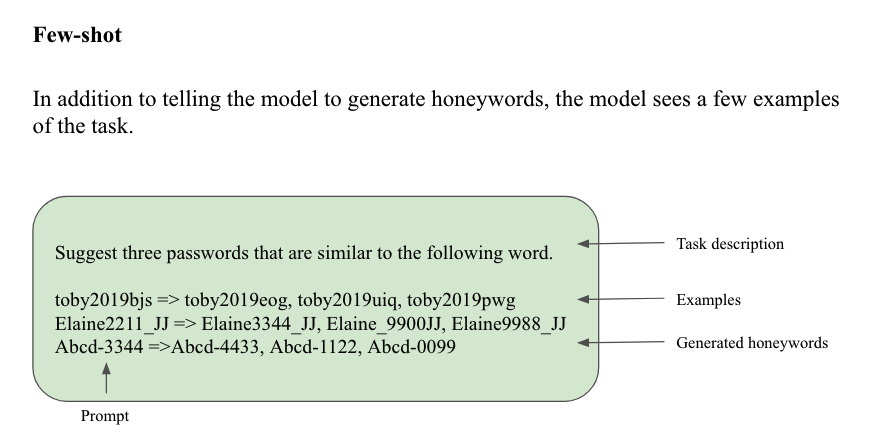}
  \caption{Give GPT-3 two examples for generating better honeywords. Through examples, the model can learn the expected pattern of honeywords and create honeywords with the same pattern as the real password.}
  \label{fig:few-shot}
\end{subfigure}
\caption{The comparison of honeywords generated by GPT-3 in two conditions: given no examples and given two examples. The model is able to generate honeywords that conform to the operators' expectations by given demonstrations.}
\label{fig:shot}
\end{figure*}

\textbf{The temperature.} The temperature is a numeric variable between 0 and 1 that effectively regulates the model's degree of confidence when generating predictions. A lower temperature implies that the model will take fewer risks, and the honeywords created will be more precise and predictable. While increasing the temperature results in more diversified honeywords, a drawback is it may also cause the PII in genuine passwords to be replaced with other words. Table \ref{tab:temp} contains examples of honeywords formed at temperatures 0 and 1.

\textbf{Zero-shot learning and Few-shot learning.} Zero-shot learning refers to a situation in which no demonstrations are permitted and the model is given simply a plain language description of the task. In comparison, few-shot learning refers to a situation in which the model is given a few demonstrations of the task during inference time, but the model is not re-trained on the demonstrations. This is particularly advantageous since many websites have varying policies regarding password creation, such as begining with letters and requiring uppercase, lowercase, symbols, and numbers. When the operators demonstrate how they want the honeywords to appear, GPT-3 will generate honeywords that match the examples. An image illustration is in Figure \ref{fig:shot}.

\subsection{How Operators can use GPT-3 for targeted honeyword generation} 
Integration of GPT-3 into the operator's system for honeyword production is relatively simple.  Unlike other HGTs that require
to be trained on a leaked password dataset \cite{dionysiou2021honeygen}, our HGT only requires the operator to simply incorporate the OpenAI API \cite{openai}, and give a specific prompt and temperature in the  $GEN(k, PW_i$) command and no training on leaked password is required.

\section{Evaluation}
We evaluated our HGT by conducting human studies and wanted to validate our hypothesis: When a victim's username is provided, attackers need more attempts to correctly find the victim's real password when honeywords are generated by GPT-3 than the control condition.

For the HGT in the control condition, we use the chaffing-by-tweaking HGT which was initially presented in \cite{juels2013honeywords} and mainly relies on random letter, digit, and symbol substitution. We choose to use chaffing-by-tweaking instead of other recently proposed methods in the literature because other methods are more vulnerable to targeted attacks, with a typical example mentioned in Section \ref{targeted}, and some more examples shown in Table \ref{tab:fasttext}. Dionysiou et al. \cite{dionysiou2021honeygen} highlight the intricacy of developing tweaking rules in such a way that it could be difficult for an attacker to distinguish the password from its changed versions. For example, if a chaffing-by-tweaking strategy randomly perturbs the last three characters of a password, the adversary may easily conclude that the authentic password is the first one in the instances ``18!morning'', ``18!morniey'', and ``18!gorndge''. Thus, they replace all occurrences of a particular symbol in a given password with a randomly chosen alternate symbol, lower-case each letter in a password with probability $p=0.3$, upper-case each letter in a password with probability $f=0.03$, and replace each digit occurrence with probability $q=0.05$. \cite{dionysiou2021honeygen} contains the pseudocode and rationale for the assignment of $p, q$, and $f$. A few examples of honeywords developed by tweaking are included in Table \ref{tab:tweak}. Table \ref{tab:GPTHW} shows some honeyword samples generated by GPT-3 for comparison.
\begin{table}[]
\centering
\caption{Honeyword samples generated by \textit{fasttext}. The word embedding machine learning model \textit{fasttext} has been trained on a subset of the \textit{rockyou} dataset. Generated honeywords usually do not contain any current-user-specific PII.}
\label{tab:fasttext}
\begin{tabular}{cccc}
\hline
\textbf{Passwords}                                                                     & \textbf{deshaun96} & \textbf{dafnny\_24} & \textbf{toby2009bjs} \\ 
\hline
\multirow{4}{*}{\textbf{\begin{tabular}[c]{@{}c@{}}Honeywords\\ by \textit{fasttext}\end{tabular}}} & foodlion21         & snuffy22            & yaiy236           \\
 & cutechica1 & octavia3  & cooneoos3a \\
 & felli1330  & Bushido07 & kmt3299 \\
 & boedha21   & Dampire2  & broloond \\ 
\hline
\end{tabular}%
\end{table}
\begin{table}[]
\centering
\caption{Honeyword samples generated by tweaking.}
\label{tab:tweak}
\begin{tabular}{cccc}
\hline
\textbf{Passwords}                                                                     & \textbf{deshaun96} & \textbf{dafnny\_24} & \textbf{toby2009bjs} \\ 
\hline
\multirow{4}{*}{\textbf{\begin{tabular}[c]{@{}c@{}}Chaffed\\ Honeywords\end{tabular}}} & DeShauN37         & dafnny=96            & toBy2009Bjs          \\
 & deshaun87  & dafNnY$<$44  & tOby2010bjS \\
 & deshAUn66  & dAfnny+47 & tobY1009bjs \\
 & DesHaun56    & Dafnny\@75  & TOby3509bJs \\ 
\hline
\end{tabular}%
\end{table}
\begin{table}[]
\centering
\caption{Honeyword samples generated by GPT-3. The model does not need to be fine-tuned on any dataset, giving it a proper prompt and temperature is enough to generate high-quality honeywords. The real passwords are from the \textit{rockyou} dataset. The honeywords generated are in line with people's password creation behavior: making minor variations on existing passwords while keeping the tokens with semantic meaning intact.}
\label{tab:GPTHW}
\begin{tabular}{cccc}
\hline
\textbf{Passwords}                                                                     & \textbf{deshaun96} & \textbf{dafnny\_24} & \textbf{toby2009bjs} \\ 
\hline
\multirow{4}{*}{\textbf{\begin{tabular}[c]{@{}c@{}}Honeywords \\by GPT-3\end{tabular}}} & deshaun97         & dafnny\_25           & toby2009bjd          \\
 & deshaun98 & dafnny\_28  & toby2009bjx \\
 & deshaun02  & dafnny\_29 & toby2009bjz \\
 & deshaun07   & dafnny\_23  & toby2009bjh \\ 
\hline
\end{tabular}%
\end{table}

\subsection{Experiment Design}
We conducted a pilot study and asked two of the researchers involved in this work to answer a survey containing questions for both HGTs. Our study did not require research ethics approval since it only involved authors of this research work and it was non-intrusive with no potential risks identified.
In our experiment, we have one independent variable: the type of HGT; two conditions: GPT-3 and tweaking; and one dependent variable: the number of attempts required to find the real password.

Participants were required to answer 12 rank-order questions, which match 6 sets of honeyword samples produced from each of the two HGTs. Each
question has 19 honeywords and 1 real password belonging to different users. The order of the 20 sweetwords were randomized. The participants were asked to sort the 20 sweetwords in each question according to their level of confidence that the sweetword is a real password, and the user's username is provided. At the end of the survey, we asked participants how hard the task is, and there are five choices ranging from  ``not hard at all'' to  ``extremely hard'', as shown in Appendix \ref{sentimental}.

\textbf{Dataset.} 
We generated honeywords based on real passwords from a leaked compilation of various password breaches over time. The dataset was first discovered by 4iQ in the Dark Web\footnote{1.4 Billion Clear Text Credentials Discovered in a Single Database: https://medium.com/4iqdelvedeep/1-4-billion-clear-text-credentials-discovered-in-a-single-database-3131d0a1ae14}. The dataset consists of 1.4 billion email-password pairs, with 1.1 billion unique emails and 463 million unique passwords. Duplicate email-password pairs were removed by an unknown curator. The listed leaks are from websites such as Canva, Chegg, Dropbox, LinkedIn, Yahoo!, Poshmark, etc. We intentionally chose the email-password pairs where the password includes information from the email as our real password candidates. In the real world scenario, the password may not contain information in the email address, but it may contain other PII, and we assume that the attacker could have access to these PII. Due to the constraint of the data we have, we only use email address as the PII in the experiments. To alleviate participants' cognitive burden, we extract the part before the symbol ``@'' as each user's username and provide the username instead of email address to participants. 

To mitigate the negative impacts of learning effects and fatigue caused by the within-group experiment, we employed the Balanced Latin Square Design
\cite{bradley1958complete}, in which each HGT appears the same amount of times as the first and second. Table \ref{tab:latinsquare} illustrates the sequence in which each HGT appears in the survey’s 12 questions.

\begin{table}[]
\centering
\caption{The order of each HGT appearing in the 12 survey questions.}
\label{tab:latinsquare}
\begin{tabular}{|c|c|}
\hline
GPT-3 & Tweaking \\ \hline
Tweaking & GPT-3 \\ \hline
GPT-3 & Tweaking \\ \hline
Tweaking & GPT-3 \\ \hline
GPT-3 & Tweaking \\ \hline
Tweaking & GPT-3 \\ \hline
\end{tabular}%
\end{table}

As discussed in Section 3.1, the prompt, temperature and given examples can affect the quality of GPT-3 generated honeywords. For the pilot experiment, we choose to use a temperature of 0.65, and give the model instruction of ``Suggest 19 distinct passwords that are similar to \uppercase {real password}, and are passwords that a LinkedIn user with username \uppercase{username} would use.''  

\subsection{Results}
Our analysis is based on the responses to our survey that each participant provided. We want to determine if there is a significant difference in the average number of attempts required for users to properly guess the real password in the GPT-3 condition compared with the control condition.

We concatenated the responses for each HGT and got a dataset containing two columns (the two HGTs), and 12 ($6\times2$) rows, where each value represents the attempts needed to find the real password in one of the questions in the corresponding HGT. We analyzed the data using a paired-samples t-test to examine if there are significant difference between attempts required to find the real password for GPT-3 vs tweaking. As a result, due to the sample size being too small, we did not find a significant difference between the attempts required to find the real passwords in the two conditions when users' username is provided. We anticipate that if we used a large enough sample size, a significant difference would be observed between the two conditions. Nonetheless, both participants expressed that the task was extremely difficult to complete, and they took an average of 17 minutes and  7 seconds to complete all 12 questions. A sample question in the survey is provided in Appendix \ref{sample}. 

\section{Discussion}
We talk about the limitations of our study and future directions in this section.

\textbf{Quantitative Evaluation.} Targeted online password attack is an underestimated concern compared with trawling attacks \cite{wang2016targeted}. We are the second to work on targeted honeyword generation techniques, and the first to use language models, notably GPT-3, to produce honeywords. We are highly confident about the novelty and significance of our methodology. However, because there are too few works being done in this field, we do not have proper benchmarks in targeted attacks that can be used for our evaluation process. An attack model is required to generate the commonly used metrics, namely the flatness and success-number graphs \cite{wang2018security}, but there is no generalized attack model that can be used for targeted purposes; the most popular Normalized Top-PW attack model is for trawling attacks \cite{9799343} in which the attackers are assumed to know nothing about the victims. We recognize that our technique lacks quantitative assessment, and we are developing quantitative analysis to demonstrate the resilience of our HGT against targeted attacks. We also intend to draw the community's attention to targeted scenarios, since trawling situations have been intensively studied, but targeted honeyword generation and attack models are under-researched yet a pressing problem, as outlined in Section 1.1.


\textbf{Qualitative Evaluation}. The pilot study shows distinguishing between the real password and honeywords is an extremly tough task, although there is not a significant difference between the GPT-3 and tweaking HGTs since we conducted experiments only with two participants. We plan to conduct a human study with a larger sample size via crowd-sourcing platforms, and we speculate that a larger sample size will result in a significant difference between the two HGTs.

\textbf{Irreversibility.} The irreversibility of a HGT is critical. We need to make sure that even when attackers know the prompt and the temperature we were using for generating honeywords, they still cannot reproduce the honeywords we generated. This is ensured by careful prompt-engineering \cite{kojima2022large,reynolds2021prompt} and temperature setting. We suggest to set temperature to 1 to get the most randomness \cite{brown2020language}, and after experimenting with various prompts, we decided to use the prompt "Derive 19 words that are similar to $real password$, and contain the word $PII$. The length of the words should be more than 10." since it generates the most diversified honeywords compared with other prompts we experimented with, and the honeywords generated each time is different.    

\section{Conclusions}
In this paper, we propose a novel HGT which utilizes GPT-3 to generate high-quality honeywords that contain PII existing in users' real passwords. Honeywords generated by GPT-3 are robust to targeted attacks where attackers get access to both breached password databases and users' personal identifiable information. Unlike other machine learning-based HGTs, GPT-3 can be easily integrated into any current password-based authentication system without any further training on real passwords. Additionally, we compared GPT-3's performance to a tweaking technique by conducting a human study with two researchers. It is proved that to find the real password among all sweetwords is an extremely difficult task and a larger sample size may be required to show a significant difference between the two HGTs.

\section*{Acknowledgement}
The authors acknowledge the support of the Natural Sciences and Engineering Research Council of Canada (NSERC), funding reference number RGPIN-2018-05919.

\bibliographystyle{splncs04}
\bibliography{References}
\end{document}